\documentclass[conference]{IEEEtran}
\IEEEoverridecommandlockouts
\usepackage{cite}
\usepackage{amsmath,amssymb,amsfonts}
\usepackage{algorithmic}
\usepackage{graphicx}
\usepackage{textcomp}
\usepackage{xcolor}
\usepackage[bookmarks=true]{hyperref}

\usepackage[caption=false, font=footnotesize]{subfig}
\usepackage{graphicx}
\usepackage{xspace}
\usepackage{siunitx}
\usepackage{capt-of}
\usepackage{adjustbox}
\usepackage[compact]{titlesec}
\usepackage{siunitx}
\renewcommand{\baselinestretch}{0.979} 

\def\BibTeX{{\rm B\kern-.05em{\sc i\kern-.025em b}\kern-.08em
    T\kern-.1667em\lower.7ex\hbox{E}\kern-.125emX}}
\begin{document}

\newcommand{\nanodrone}{nano quadcopter\xspace}
\newcommand{\nanodrones}{nano quadcopter\xspace}
\newcommand{\drl}{Deep RL\xspace}
\newcommand{\fixme}[1]{\textcolor{red}{#1}}
\newcommand{\Fig}[1]{Figure~\ref{#1}}
\newcommand{\airl}{Air Learning\xspace}
\title{Learning to Seek: Autonomous\\ Source Seeking with Deep  Reinforcement\\ Learning Onboard a Nano Drone Microcontroller
}


\author{Bardienus P. Duisterhof$^{1,3}$~~~Srivatsan Krishnan$^{1}$~~~Jonathan J. Cruz$^{1}$~~~Colby R. Banbury$^{1}$~~~William Fu$^{1}$ \\\\ Aleksandra Faust$^{2}$~~~Guido C. H. E. de Croon$^{3}$~~~Vijay Janapa Reddi$^{1}$
\thanks{$^{1}$Harvard University, $^{2}$ Robotics at Google, $^{3}$Delft University of Technology - bduisterhof@g.harvard.edu. The work was done while Bart was a visiting student at Harvard.}}

\maketitle

\begin{abstract}

We present fully autonomous source seeking onboard a highly constrained nano quadcopter, by contributing application-specific system and observation feature design to enable inference of a deep-RL policy onboard a nano quadcopter. Our deep-RL algorithm finds a high-performance solution to a challenging problem, even in presence of high noise levels and generalizes across real and simulation environments with different obstacle configurations. 
We verify our approach with simulation and in-field testing on a Bitcraze CrazyFlie using only the cheap and ubiquitous Cortex-M4 microcontroller unit. The results show that by end-to-end application-specific system design, our contribution consumes almost three times less additional power, as compared to competing learning-based navigation approach onboard a \nanodrone. Thanks to our observation space, which we carefully design within the resource constraints, our solution achieves a 94\% success rate in cluttered and randomized test environments, as compared to the previously achieved 80\%. We also compare our strategy to a simple finite state machine (FSM), geared towards efficient exploration, and demonstrate that our policy is more robust and resilient  at obstacle avoidance as well as up to 70\% more efficient in source seeking. To this end, we contribute a cheap and lightweight end-to-end tiny robot learning (tinyRL) solution, running onboard a \nanodrone, that proves to be robust and efficient in a challenging task using limited sensory input. 
\end{abstract}

\section{Introduction}

Source seeking is an important application for search and rescue, inspection, and other jobs that are too dangerous for humans. Imagine cheap and disposable aerial robots inspecting ship hauls for leaks, aiding search for survivors in mines, or seeking a source of radiation in nuclear plants. For that reality, we need small, agile, and inexpensive robots capable of fully-autonomous navigation in GPS denied environments that can be deployed quickly, without additional set-up or training.

\begin{figure}
    \centering
    \includegraphics[width=0.8\linewidth]{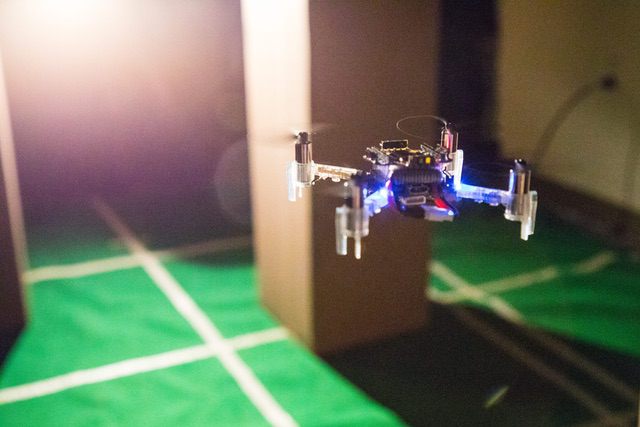}
    \caption{CrazyFlie \nanodrone running a deep reinforcement learning policy fully \emph{onboard} with robust obstacle avoidance and source seeking.}
    \label{fig:czf_config}   
\end{figure}
\begin{figure*} 
\adjustbox{valign=t}{
\begin{minipage}[t]{\linewidth}
    \centering
      \subfloat[Air Learning training env.\label{1a}]{%
        \includegraphics[width=0.215\linewidth]{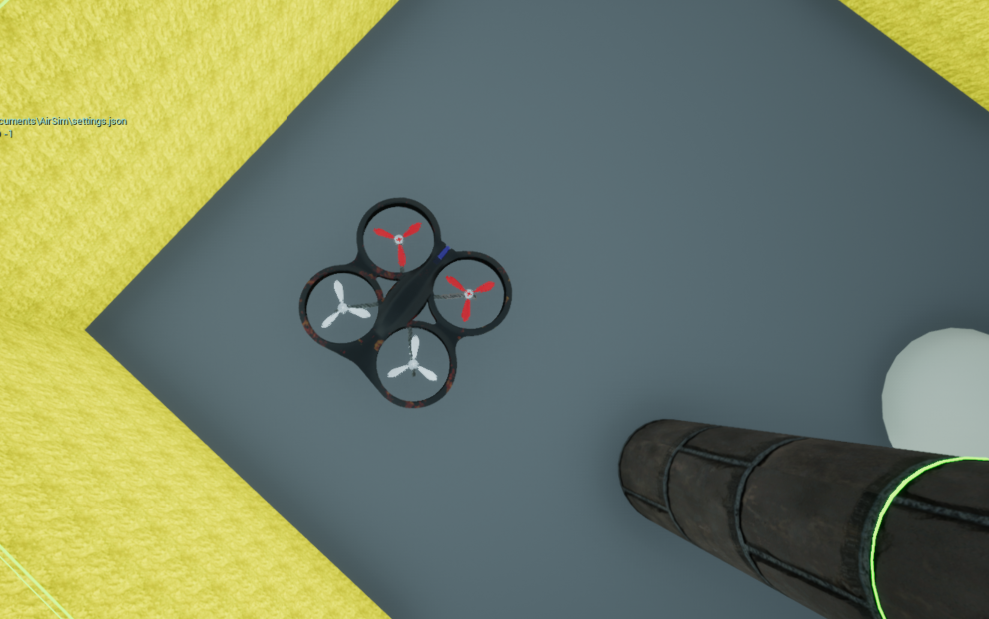}}
        \hfill
  \subfloat[0 obstacles.\label{1b}]{%
       \includegraphics[width=0.24\linewidth]{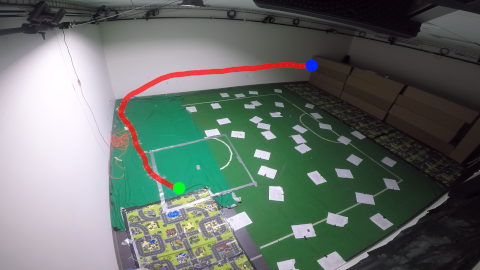}}
    \hfill
  \subfloat[3 obstacles.\label{1c}]{%
        \includegraphics[width=0.24\linewidth]{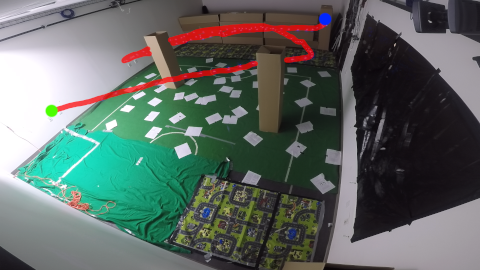}}
    \hfill
  \subfloat[7 obstacles.\label{1d}]{%
        \includegraphics[width=0.24\linewidth]{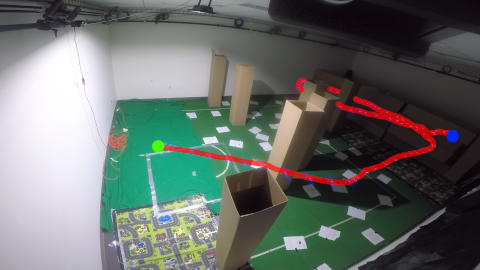}}
  \caption{Air Learning training environment (\ref{1a}), and three distinct trajectories in flight tests (\ref{1b}-\ref{1d}). Blue dot is the start and green dot is the destination.}
  \label{fig:trajectories_flight}
  \end{minipage}
  }
  \end{figure*}
Nano quadrotors are a lightweight, cheap, and agile hardware platform, and an ideal candidate for source seeking in GPS denied environments. To make them work, we need to add appropriate sensors and navigation software. However, they are severely resource constrained. Their available memory, battery, and compute power is limited. Those constraints pose challenges to the existing autonomous navigation methods, and the sensor and software selection needs to be carefully designed. The memory constraints means the system cannot store large maps used in traditional planning, the battery constraints means that we need to consider energy consumption of the system \cite{10.1109/MICRO.2018.00077}, and the limited compute power means that large neural networks cannot run. 

Source seeking applications needs motion planning capable of obstacle avoidance that can be deployed quickly, without apriori knowledge of the obstacle placement, without the need to configure or re-train for a specific condition or environment. Instead, the algorithms we deploy need to generalize to a wide range of unknown environments for effective deployment. The currently available finite state machines on nano quadcopters~\cite{SR_Kim} are designed for certain conditions and environments, and may not generalize well. Traditional navigation solutions like SLAM cannot run, as they require more memory and compute than our \nanodrone can offer. Instead, we need a mapless navigation solution that uses little sensory input, memory and compute, while also generalizing to different environments. 



To address these challenges, we present a fully autonomous source-seeking nano quadcopter (Figure~\ref{fig:czf_config}), by equipping nano UAVs with sparse light-weight sensors, and using reinforcement learning (RL) to solve the problem. We contribute POMDP formulation including source observation features, and sensor modelling. Our method operates under memory, compute, and energy constraints with zero-shot transfer from simulation to reality, and generalizes to new environments without prior training. We fit an ultra-tiny neural network with two hidden layers, which runs at 100~Hz on a commercially available off-the-shelf ARM Cortex-M4 microcontroller unit (MCU). Our source-seeking \nanodrone, shown in \Fig{fig:czf_config}, seeks a light source as a proxy for any point source, such as a radiation source. We use a light source as it allows for better understanding of behavior in real flight tests. Our \nanodrone locates the light source in challenging conditions, with varying obstacle counts and positions, even beyond the environments presented in simulation (Figure~\ref{1a}). The robot first explores, when it is far away from the source and it detects no gradient in light but just noise. It then seeks reactively for the light source when it detects gradients (Figure~\ref{fig:trajectories_flight}).

We compare our work with baselines in three areas: 1) nano quadcopter system design, 2) source seeking performance and 3) obstacle avoidance robustness. From a systems perspective, the state of the art solution for learning-based navigation onboard a nano quadcopter~\cite{dronenet} consumes almost three times more additional power. From a source seeking perspective, the state of the art RL light seeking approach~\cite{light_seeking_q-values} reaches an 80\% success rate in 'simple' simulated environments, and 30\% in 'complex' simulated environments, which we outperform by reaching a 94\% success rate in real flight tests in cluttered and randomized test environments (Figure~\ref{1b}-\ref{1d}). Finally, we compare against a finite state machine (FSM) geared towards exploration, which we outperform by more robust obstacle avoidance and efficient source seeking. Hereby, we show that our simulator-trained policy provides a high-performance and robust solution in a challenging task in on-robot experiments, even in environments beyond those presented in simulation.

\section{Related Work}

Deep reinforcement learning has proven to be a promising set of algorithms for robotics applications. The fast-moving deep reinforcement learning field is enabling more robust and accurate but also more effortless application~\cite{autorl2}.  Lower-level control has been demonstrated to be suitable to replace a rate controller~\cite{review_2_deeprl_control,koch,caltech_low_level} and was recently used to perform model-based reinforcement learning (MBRL) with a CrazyFlie~\cite{czf_deeprl_low_level}. High-level control using deep reinforcement learning for obstacle avoidance has been shown with different sets of sensory input~\cite{deeprl_hybrid_czf,airlearning}, but not yet on a \nanodrone. Although light seeking has been demonstrated before using Q-learning~\cite{light_seeking_q-values}, it used multiple light sensors and reached a success rate of 80\% in simple environments and 30\% in complex environments. Thanks to our observation space design and larger network, \emph{we present a deep reinforcement learning-based model for robust and efficient end-to-end navigation of a source seeking \nanodrone, including obstacle avoidance, that beats the current state of the art}. 

Traditional source seeking algorithms can be divided into four categories~\cite{review}: 1) gradient-based algorithms, 2) bio-inspired algorithms, 3) multi-robot algorithms, and 4) probabilistic and map-based algorithms. Even though gradient-based algorithms are easy to implement, their success in source seeking has been limited due to their unstable behavior with noisy sensors.
Previous algorithms have yielded promising results, but rarely considered obstacles~\cite{Bourne2019}. Obstacles are important as they make the problem harder, not just from an avoidance perspective, but also considering phenomena like shadows. \emph{We contribute a deep RL approach, capable of robust source seeking and obstacle avoidance on a nano drone.}

In a multi-agent setup, particle swarming~\cite{PSO,PSO_SAFE} has shown to be successful in simulation. However, PSO swarms require a positioning system, and lack laser-based obstacle avoidance. Finally, probabilistic and map-based algorithms are more flexible but require high computational cost and accurate sensory information. In contrast to traditional methods, deep reinforcement learning can learn to deal with noisy inputs~\cite{RL-noise} and effectively learn (optimal) behavior for a combination of tasks (i.e., generalize). Hence, source seeking on a \nanodrone is a suitable task for deep reinforcement learning, as it can combine obstacle avoidance with source seeking and deal with extraordinary noise levels in all sensors. \emph{We leverage these advantages of deep reinforcement learning to produce a robust and efficient algorithm for mapless navigation for source seeking.}
	

\section{Method}
\label{sec:result}
We present our application-specific system design (Section~\ref{sec:crazyflie}), using lightweight and cheap commodity hardware. Next, we describe our simulation design (Section~\ref{sec:simulation}), used to train deep-RL policies by randomizing the training environment and deploying a source model. Finally, in Section~\ref{sec:policies}, we show the POMDP formulation.


\subsection{System Design}
\label{sec:crazyflie}

\begin{figure} [t]
    \centering
    \centering
    \includegraphics[width=0.8\linewidth]{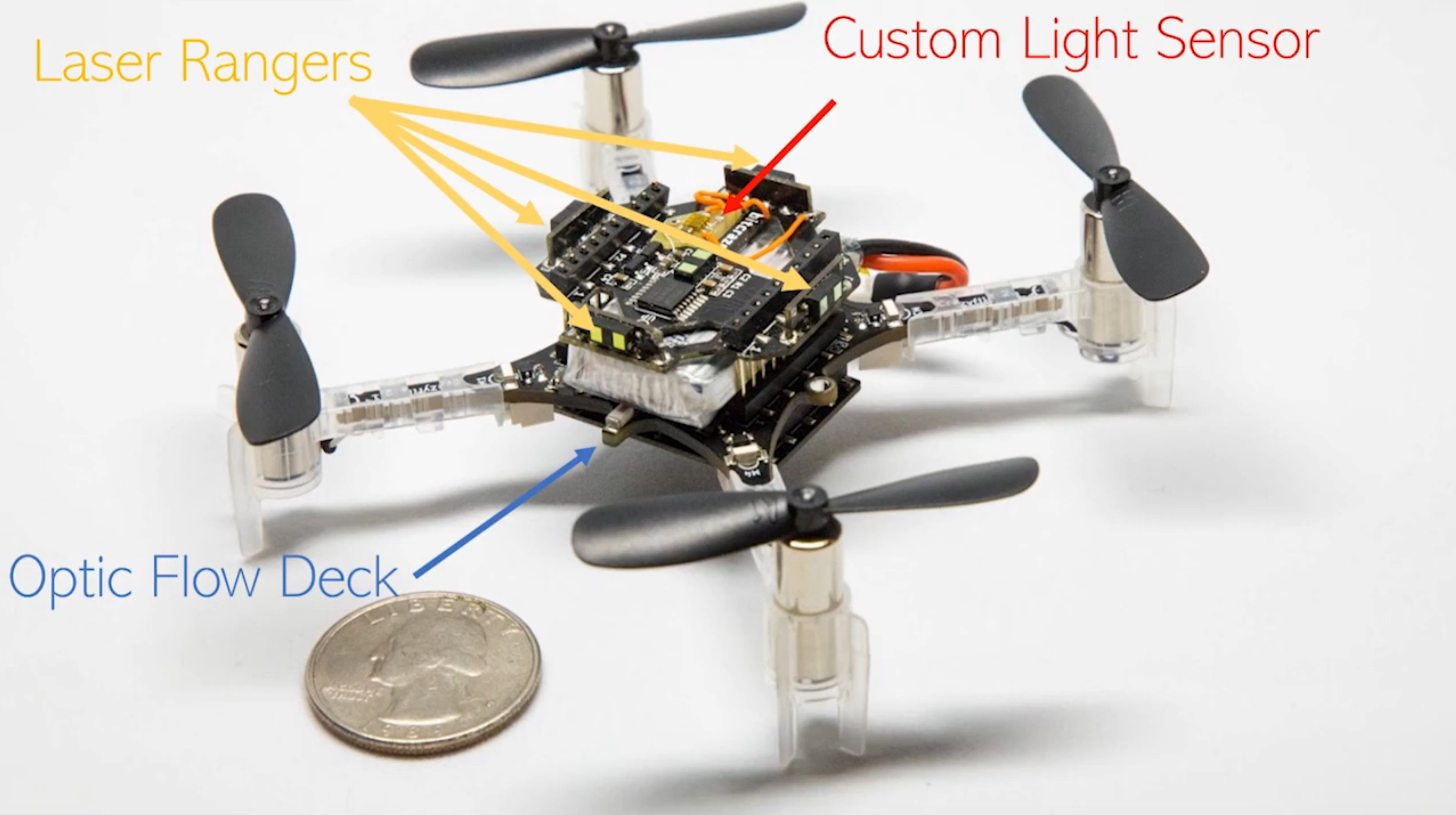}
    \caption{Our BitCraze CrazyFlie with multiranger deck, custom TSL2591 light sensor board and an optic flow deck.}
    \label{fig:czf_board}
\end{figure}

We configure a BitCraze CrazyFlie \nanodrone for source seeking, taking into account its physical and computational limitations, as visible in Table~\ref{tab:czf_specs}. While adding a camera may be useful for navigation, its added weight and cost make effective deployment more difficult. Instead, we use only cheap, lightweight and robust commodity hardware. We configure our \nanodrone with laser rangers for obstacle avoidance, an optic flow sensor for state estimation, and a custom light sensor to seek a light source (Figure~\ref{fig:czf_board}). The \nanodrone carries four laser rangers with a range of approximately \SI{5}\meter, facing in the negative and positive direction of the $x$ and $y$ axis of the robot' body frame. We attach an ultra-tiny PCB to the laser ranger board, fitting an upward-facing TSL2591 light sensor.




\renewcommand{\arraystretch}{1.1}

\begin{table}[t]
\centering

    \begin{tabular}{|l|l|l|S[table-format=3.2]|}
    \hline
    \textbf{Developer}        & \textbf{Bitcraze}        & \textbf{Parrot}      & \textbf{Delta}                   \\ \hline\hline
    \textbf{Vehicle}          & CrazyFlie 2.1   & Bebop~2                       & \\ \hline
    \textbf{Takeoff weight}   & \SI{27}\gram             & \SI{500}\gram        & 18.5x                     \\ \hline
    \textbf{Max payload}      & \SI{15}\gram            & \SI{70}\gram          & 4.6x                   \\ \hline
    \textbf{Battery (LiPo)}   & 250 mAh         & 2700 mAh                      & 10.8x     \\ \hline
    \textbf{Flight time}       & \SI{7}\minute          & 2\SI{5}\minute        & 3.6x                    \\ \hline
    \textbf{Size (WxHxD)}     & 9.2 cm x 9.2 cm     & 32.8 cm x 38.2 cm         & 12.7x           \\ \hline
    \end{tabular}%
    \vspace{2mm}
        \caption{CrazyFlie vs. Bebop2. The delta between them is significant.}
    \label{tab:czf_specs}

\end{table}

\subsection{Simulation Environment}
\label{sec:simulation}

\begin{figure}
    \centering
    \includegraphics[width =0.8\linewidth]{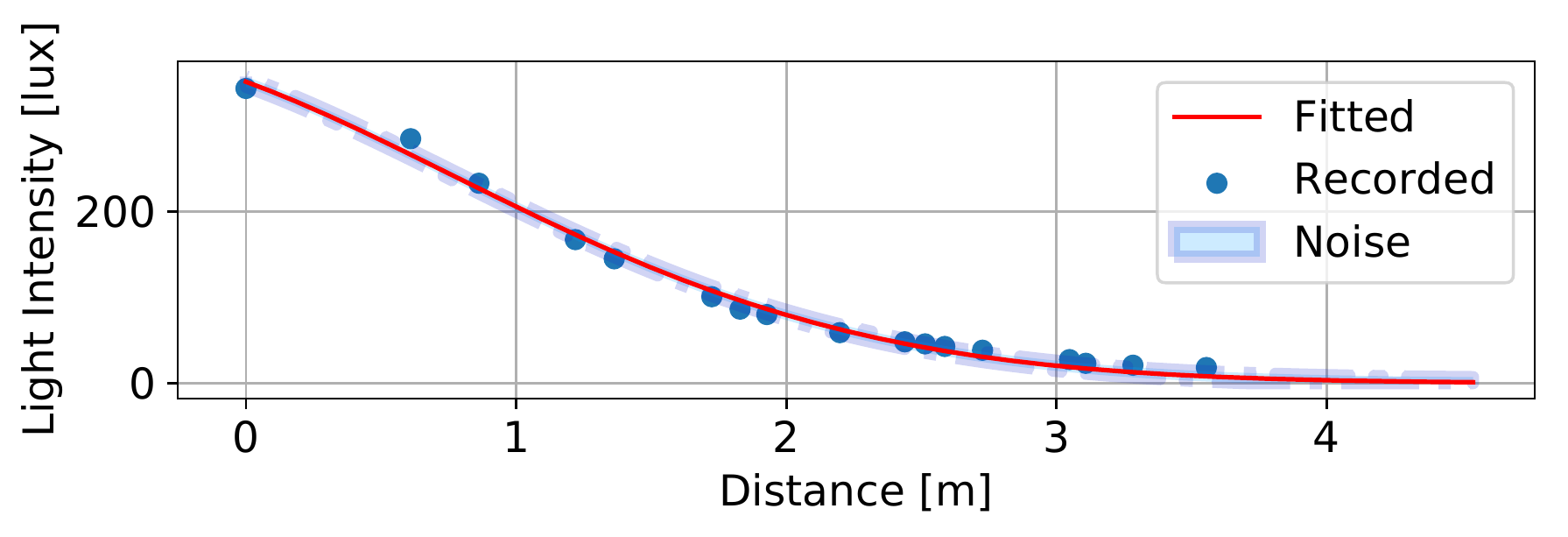}
    \caption{Light intensity describing the function as used in training with 3$\sigma$ (standard deviations) of the noise.}
    \label{fig:light_fit}

\end{figure}

Source modelling is key to the success of our \nanodrone. We model source intensity as a function of the distance from the source. We generate this function by capturing data in the testing environment with the light source present. We capture the light intensity in a two-dimensional grid with our light sensor. Once captured, we use the data to fit a function with two requirements: 1) $\lim_{dist\to0} <  \infty$ and 2) $\lim_{dist\to \infty} = 0$. A Gaussian function meets both requirements and is shown in Figure~\ref{fig:light_fit}. The function has the form: $f(x) = a\cdot e ^{-\frac{(x-b)^2}{2c^2}}$ with $a = 399.0,b = -2.6,c =5.1$.
The R-squared error, measuring the goodness-of-fit, is 0.007, implying a high-quality fit.
Additionally, we inject Gaussian noise with a standard deviation of 4.
The noise observed in recordings had a standard deviation of 2; however, in flight with unstable attitude, we expect more noise, so we inject more noise than recorded. In flight tests (Section~\ref{sec:flight_eval}), we present the robustness of this function when shadows and reflections are present.


    

\subsection{POMDP Setup}
\label{sec:policies}

To learn to seek a point source, we choose reinforcement learning with partial state observations, which we model as a Partially Observable Markov Decision Process (POMDP). The agents is modeled as a POMDP by the tuple $\left( O, A, D, R, \gamma \right)$ with continuous observations and discrete actions. We train and deploy a DQN~\cite{DQN} algorithm, using a feedforward network with two hidden layers of 20 nodes each and  activations. 

The observations, $\mathbf{o}=\left(l_1,l_2,l_3,l_4,s_1,s_2\right) \in O$, consist of four laser ranger values in front/right/back/left directions , $l_1$-$l_4$,, and two custom `source terms', $s_1$ and $s_2$. The source terms are inspired by~\cite{guido}, and provide an estimate for source gradient and strength. We first compute a normalized version of the light sensor readings $c$, as sensor readings are dependent on sensor settings (e.g., integration time, gain). We then add a low-pass filter and compute $c_f$ (Equation~\ref{eq:c_f}). We then compute $s_1$ (Equation~\ref{eq:s1}), which is effectively a normalized and low-pass version of the gradient of $c$ (i.e., it is the light gradient over time). Finally, we compute $s_2$ (Equation~\ref{eq:s2}), a transformation of $c_f$. 

\resizebox{.95\columnwidth}{!}{%
       
    \begin{minipage}[t]{0.4\linewidth}
        \begin{equation}
            c_{f} \leftarrow 0.9\cdot c_{f} + 0.1 \cdot c
            \label{eq:c_f}
        \end{equation}
    \end{minipage}
    \begin{minipage}[t]{0.3\linewidth}
    \begin{equation}
        s_1 = \frac{c-c_{f}}{c_{f}}
        \label{eq:s1}
        \end{equation}
    \end{minipage}
    \begin{minipage}[t]{0.30\linewidth}
        \begin{equation}
        s_2 = 2\cdot c_{f}-1
        \label{eq:s2}
        \end{equation}
    \end{minipage}
    }
    
Figure~\ref{fig:source_traces} shows traces of all variables onboard a source-seeking \nanodrone. Term $s_1$ is effectively a normalized and low-pass version of the gradient of $c$ (i.e., it is the light gradient over time). The low-pass filter has a high cutoff frequency, but is useful in filtering outliers. Normalization is crucial when far away from the source, when the gradient is small and finding the source is hard. Term $s_2$ is a transformation of $c_f$. The signal provides the agent with a proxy of distance to the source, which it can use to alter behavior. 

We have trained policies with non-filtered gradient inputs, leading to poor results (30\% succes rate). While the direct gradient only contains information about the last time-step, $c_f$ is the weighed sum of the entire history of measurements. It is a computationally cheap way of using history information, which is necessary as the fluctuations between two time-steps (\SI{0.01}\second) are noise-dominated.

To teach the \nanodrone to seek the source, the reward is computed at each step (instantaneous reward):

\begin{equation}
r = 1000 \cdot \alpha - 100 \cdot \beta - 20 \cdot \Delta D_{s} -1
 \label{eq:reward}
\end{equation}

Here, $\alpha$ is a binary variable whose value is `1' if the agent reaches within \SI{1.0}\meter~from the goal else its value is `0'. ~$\beta$ is a binary variable which is set to `1' if the \nanodrone collides with any obstacle or runs out of the 300 steps.\footnote{We set the maximum allowed steps in an episode as 300. This is to make sure the agent finds the source within some finite amount of steps and penalize a large number of steps.} Otherwise, $\beta$ is `0',  penalizing the agent for hitting an obstacle or not finding the source in time. $\Delta D_{s}$ is the change in distance, compared to the previous step. A negative $\Delta D_{s}$ means the agent got closer to the source, rewarding the agent to move to the source.

The agent's discrete actions space, $\mathbf{a} = \left(v_x, \dot{\psi} \right)^3 \in A$, consists of three pairs of target states, composed of target yaw rate and the target forward velocity ($v_x$). The target states are then realized by the low-level PID controllers. The three actions are: move \texttt{forward}, rotating \texttt{left} or \texttt{right}. The forward-moving speed is \SI{0.5}{\meter/\second}, and the yaw rate is 54$^{\circ}$/s in either direction.  Finally, the dynamics $D$ of the environment are a simple drone model developed by AirSim~\cite{airsim} and $\gamma$ was set to be 0.99.



\section{Implementation Details}
\label{sec:implementation}

We discuss the implementation details of our simulation environment (Section~\ref{sec:sim_imp}), and inference of the deep-RL policy onboard the \nanodrone (Section~\ref{sec:inference_imp})

\subsection{Simulation}
\label{sec:sim_imp}
We simulate an arena to train an agent to seek a point source. 
The agent (i.e., drone) is initialized in the middle of the room, and the point source is spawned at a random location. By randomizing the source position and obstacle positions, we arrive at a policy that generalizes to different environments with different obstacle configurations. We use the Air Learning platform~\cite{airlearning}, which couples with Microsoft AirSim~\cite{airsim} to provide a deep reinforcement learning back end. It generates a variety of environments with randomly varying obstacle density, materials, and textures. 

\subsection{Inference}
\label{sec:inference_imp}

 The CrazyFlie\cite{bitcraze_crazyflie} is heavily constrained, carrying an STM32F405 MCU. Though our MCU is constrained in memory and compute, it is widely deployed. In fact, more than 30 billion general-purpose MCUs are shipped every year.\footnote{From \href{https://www.icinsights.com/data/articles/documents/1190.pdf}{IC Insights, Research Bulletin.}} Their ubiquity makes them cheap, easy to use and expendable, and so suitable for cheap and disposable search and rescue robots.





\newcommand{\tfmicro}{TF-Lite\xspace}

Our implementation consists of a custom lightweight C library, capable of performing the necessary inference operations. The advantage of this approach is its small memory footprint and overhead, compared to a general inference framework like TensorFlow Lite for Microcontrollers~\cite{david2020tensorflow}. 

The Crazyflie has 1 MB of flash storage. The stock software stack occupies 192~kB of the available storage, while the custom source seeking stack takes up an additional 6~kB. So the total flash storage used is 198~kB, which leaves an ample amount of free storage space (over 75\%).

\begin{figure}
    \centering
    \includegraphics[trim=0 0 0 -10, clip, width=0.8\linewidth]{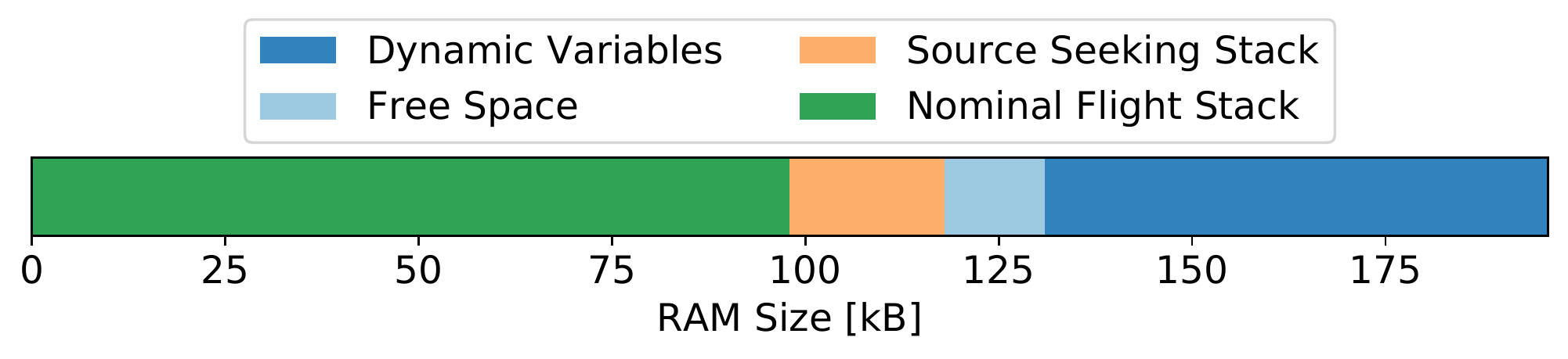}
    \caption{RAM usage on the Bitcraze CrazyFlie, using a custom float inference stack. Total free space: 12.5 kB }
    \label{fig:ram_czf_float}
\end{figure}

However, the memory constraints are much more severe. RAM availability during execution is shown in Figure~\ref{fig:ram_czf_float}. Of the 196~kB of RAM available on the Cortex-M4 microcontroller, only 131~kB is available for static allocation at compile time. The rest is reserved for dynamic variables (i.e., heap). During normal operation, the Bitcraze software stack uses 98~kB of RAM, leaving only 33~kB available for our purposes. The entire source seeking stack takes up 20.5 kB, leaving 12.5 kB of free static memory. Our policy runs at \SI{100}\hertz~in flight.


\section{Results}
\label{sec:results}

We evaluate simulation and flight results of our models. We evaluate training results (Section~\ref{sec:sim_eval_training}), introduce our baselines (Section~\ref{sec:baseline}) and evaluate the models in simulation (Section~\ref{sec:sim_inference}) and flight tests (Section~\ref{sec:flight_eval}). Finally, we analyze robot behavior (Section~\ref{sec:bahavior}) and endurance and power consumption (Section~\ref{sec:endurance}).

\subsection{Training in Simulation}
\label{sec:sim_eval_training}
\begin{figure}
        \centering
  \subfloat[Training success rates.  \label{3a}]{%
       \includegraphics[width=0.49\linewidth]{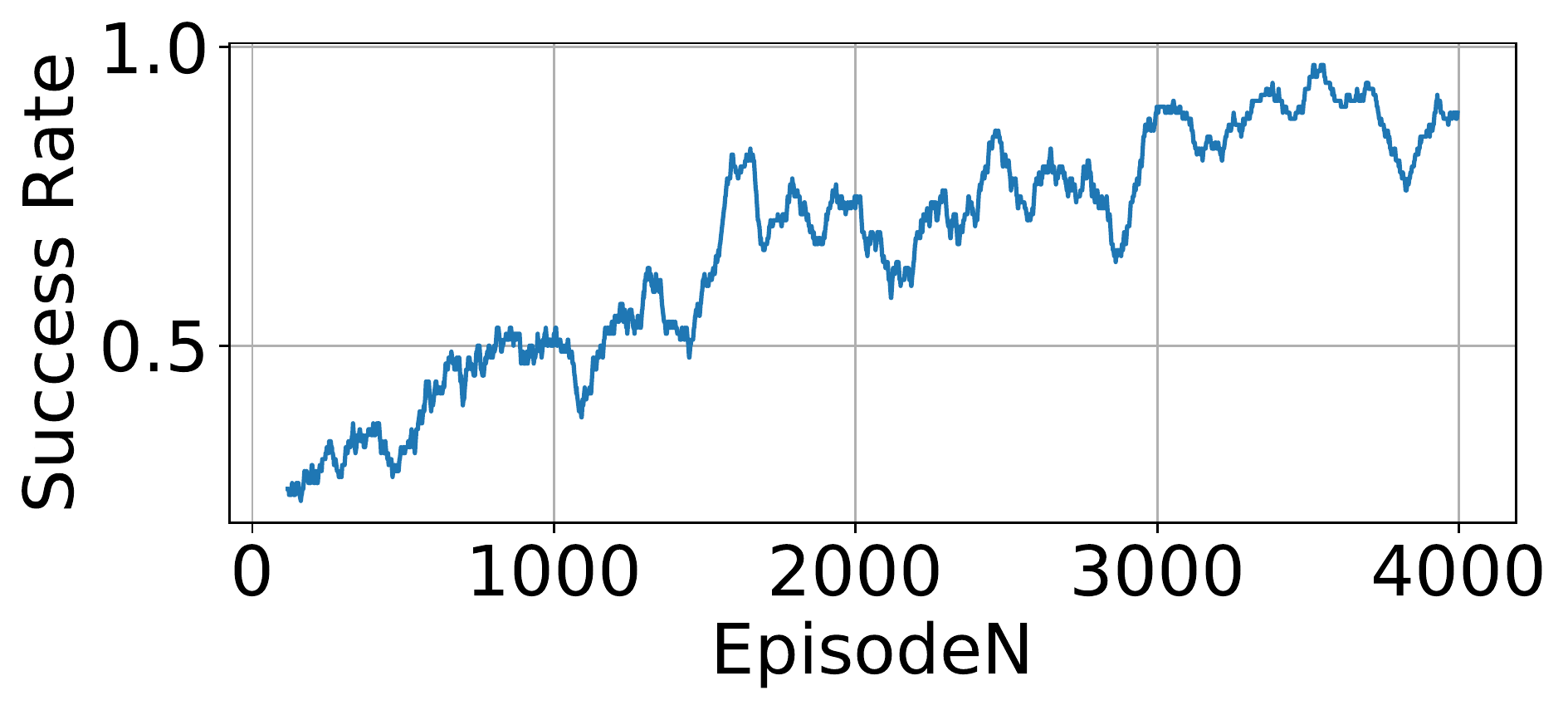}}
    \hfill
  \subfloat[ Steps to locate source.\label{3b}]{%
        \includegraphics[width=0.49\linewidth]{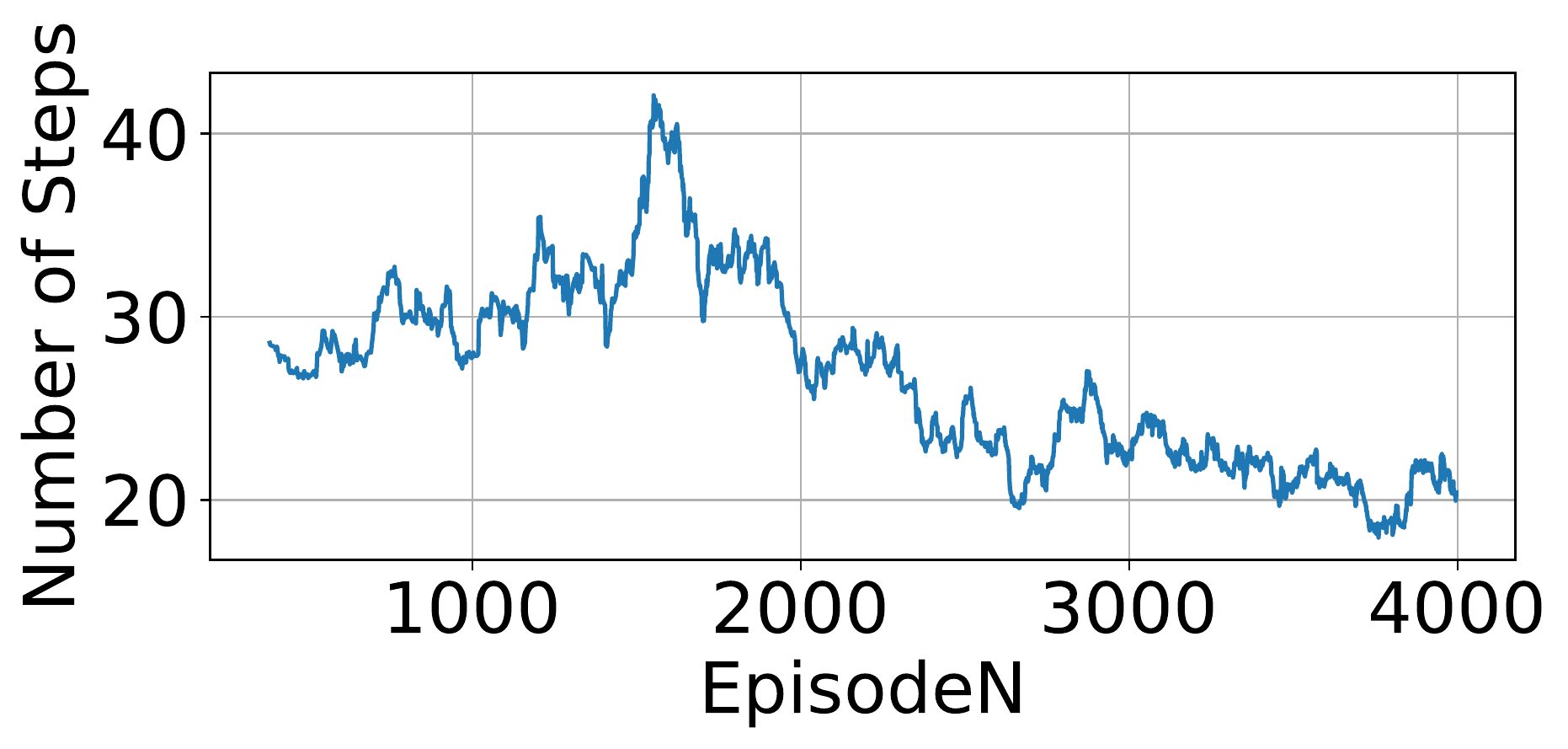}}
    \captionof{figure}{Quality metrics during training.}
  \label{fig:training}

\end{figure}

\begin{table}
\begin{tabular}{|l|c|c|c|c|}
\hline
\textbf{Model description}          & \textbf{Success} & \textbf{\# of Steps} & \textbf{Distance [m]} & \textbf{SPL} \\ \hline \hline
Our deep RL algorithm                   & 96\%         & 30.51          & 4.21   & 0.37                       \\ \hline
FSM baseline            & 84\%         & 87.73            & 10.40    &0.16                     \\ \hline
Random actions         & 30\%         & 42.13           & 3.55      &0.13                     \\ \hline
\end{tabular}
\vspace{10pt}
\caption{Models in simulation.}
\label{tab:model_eval}
\end{table}




To evaluate the learning process, we present quality metrics (success rate and number of steps) during training. As shown in Figure~\ref{fig:training}, we train our policy up to convergence at around 3,600 episodes (or 100,000 steps). A consistent upward trend in success rate is visible, while number of steps shows a consistent decrease after an initial spike. The initial spike is caused by the agent becoming more successful, hence reaching more distant targets, instead of only finding close targets. After continued training success rate quickly drops, i.e., it over-trains after around 3600 episodes. We continued training to over 8,000 episodes and never saw an improvement in performance.

\subsection{Baseline Comparison}
\label{sec:baseline}

We compare our approach with three baselines: 1) \nanodrone system design, 2) source seeking performance and 3) obstacle avoidance robustness. From a systems perspective, the state of the art solution for learning-based navigation onboard a \nanodrone~\cite{dronenet} adds almost three times more power consumption (Section~\ref{sec:endurance}), showing end to end application-specific system design can benefit mission metrics. Thanks to our low-dimensional sensory input, we perform inference at \SI{100}\hertz~on the stock microcontroller processor. 


From a source seeking perspective, the state of the art RL light seeking approach~\cite{light_seeking_q-values} reaches an 80\% success rate in 'simple' simulated environments, and 30\% in 'complex' simulated environments, which we beat by reaching a 94\% success rate in real flight tests (Section~\ref{sec:flight_eval}) in cluttered and randomized test environments. Our design of the observation space (i.e., the network inputs) has been critical in our success.

Finally, we compare against a finite state machine (FSM) geared towards exploration~\cite{baseline}, which we outperform by more robust obstacle avoidance and more efficient source seeking, as shown in Table~\ref{tab:model_eval}. As demonstrated in~\cite{baseline}, this approach is effective in exploration and often used in autonomous cleaning robots. This baseline serves to understand the difficulty of obstacle avoidance using limited sensory input, and to show our policy efficiently uses light information. We test it in the same simulation and real test environments, to put our approach into perspective. As a final baseline, we have also tested random actions, to show the effectiveness of the finite state machine (FSM). We cannot test the approach from~\cite{light_seeking_q-values}, as, to the best of our knowledge, no public code is available.

\subsection{Inference in Simulation}
\label{sec:sim_inference}


\begin{figure}

        \centering
        \includegraphics[width=0.8\linewidth]{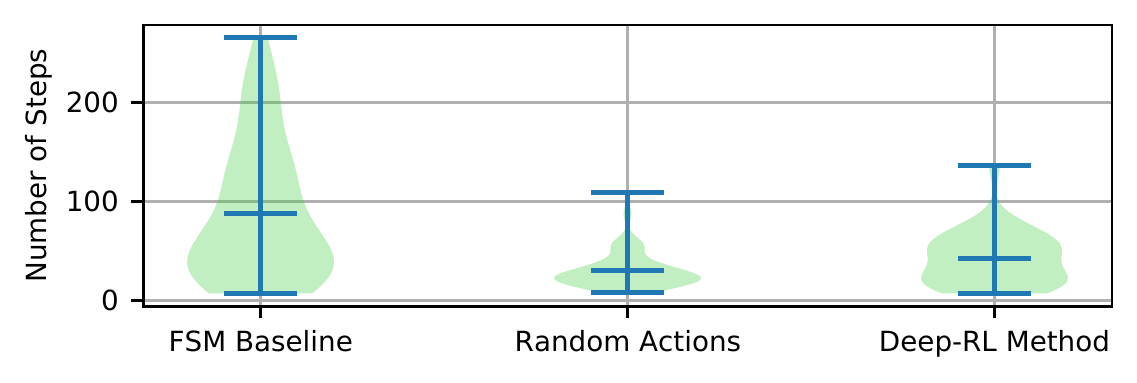}
        \caption{Number of steps in simulation over 100 runs of each algorithm.}
        \label{fig:sim_steps}
\end{figure}
\begin{figure}
        \centering
        \includegraphics[width=0.8\linewidth]{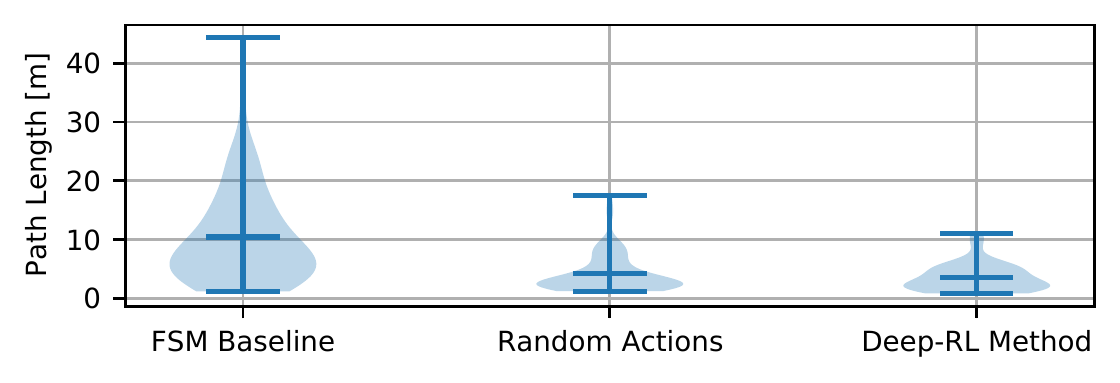}
        \caption{Path length (i.e. traveled distance) in simulation over 100 runs.}
        \label{fig:sim_distances}
\end{figure}

Training data provide limited information as the model is continuously changing. So, we evaluate our policy after training (as shown in Table~\ref{tab:model_eval}). We compare it in terms of success rate, the average number of steps, and average traveled distance. The number of steps and traveled distance are captured only when the agent succeeds. Additionally, we add the `SPL' metric: Success weighted by (normalized inverse) Path Length~\cite{anderson_spl}. We compute the SPL metric as:

\begin{equation}
    SPL = \frac{1}{N}\sum S_{i} \frac{l_{i}-1}{\text{max}(p_{i},l_{i}-1)}
\end{equation}

Here $N$ is the number of runs considered, $l_{i}$ the shortest direct path to the source, $p_{i}$ the actual flown path and $S_{i}$ a binary variable, 1 for success and 0 for failure. We subtract $l_{i}$ by 1, as the simulation is terminated when the drone is within 1 meter of the source. We do not take into account obstacles in the path, making the SPL displayed a conservative estimate.

We evaluate our finite state machine (FSM) and fully random actions in simulation. Table~\ref{tab:model_eval} and Figures~\ref{fig:sim_steps},~\ref{fig:sim_distances} show the results of testing each method for 100 runs. Our deep reinforcement learning model outperforms the FSM baseline in every metric. It finds the source in 65\% fewer steps, with a 14\% higher success rate, with 60 \% shorter paths and a 131\% higher SPL. Random actions yield shorter successful paths, as shorter paths have a higher chance of survival with random actions. In fact, the average path length over all (successful and failed) attempts for the random approach is \SI{5.7}\meter, while \SI{4.19}\meter~for our approach.

\subsection{Flight Tests}
\label{sec:flight_eval}

We use a room that is approximately \SI{5 x 5}\meter~in size (see Figure~\ref{fig:trajectories_flight}). We use a \SI{50}\watt~light source attached to the roof, radiating a $120^\circ$ beam onto the ground, as the light source.
We count a distance under \SI{0.7}\meter~as a successful run, while the drone is flying at \SI{1}{\meter /\second}~and performs inference at \SI{100}\hertz. Figure~\ref{fig:trajectories_flight} show four distinct trajectories during testing.

\begin{figure}
  \includegraphics[width=\linewidth]{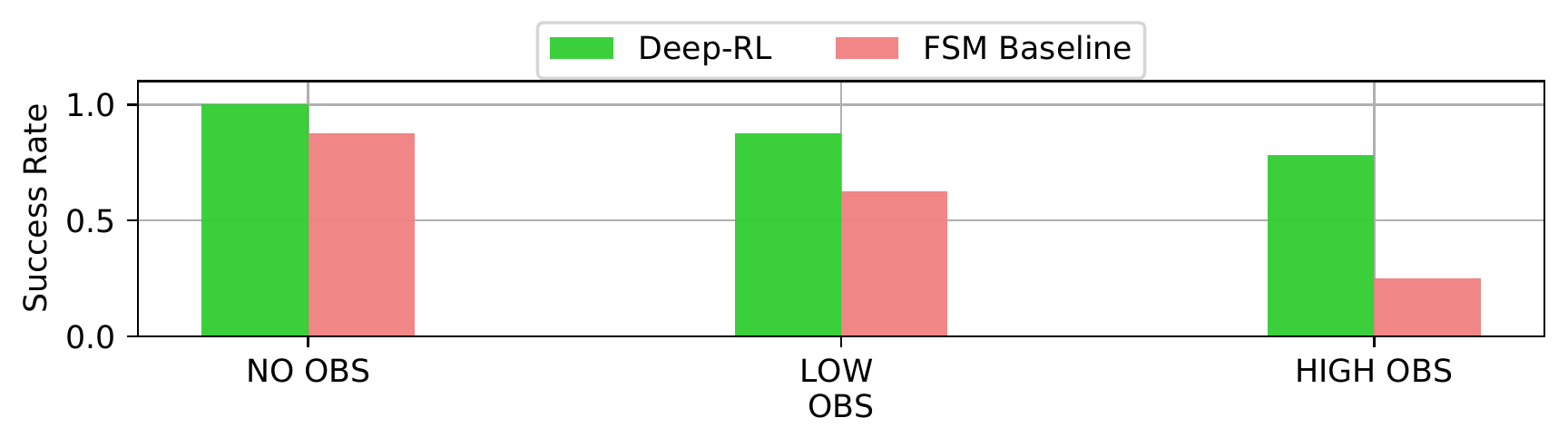}
    \caption{Success rate over 104 flight experiments, comparing our deep reinforcement learning approach with the FSM baseline. Our solution consistently performs better, especially in high-density obstacle environments.}
    \label{fig:flight_exps_success}
\end{figure}







We conduct 114 flight tests in a variety of different scenarios, involving highly cluttered environments. Across a set of representative flight tests, we get an average success rate of 94\%.
This number is measured over a set of 32 experiments, 16 with no obstacles, and 16 with 3 obstacles in the room.
This is representative to simulation as it alternates between no obstacles and sparse obstacles.
All agents were initialized at different positions, at the same \SI{4.6}\meter~from the source on one axis. On the other axis, we varied drone position between \SI{0} and \SI{4.6}\meter~from the source, resulting in an initial total source distance between \SI{4.6}\meter~and \SI{6.5}\meter.

Obstacle configuration, density, and source position were randomized. We classify three distinct obstacle densities: `NO OBS', `LOW OBS', and `HIGH OBS', featuring zero, three and seven obstacles respectively. To better understand the behavior of our algorithm, we decompose the results into two categories: 1) success rate and 2) mission time.


Over a total of 104 flight tests, we compared success rate of our model against the FSM baseline model. As can be seen in Figure~\ref{fig:flight_exps_success}, our model beats the baseline in all three obstacle density groups. The baseline reaches a 75\% success rate in a set of representative flight tests (`NO OBS' and `LOW OBS'), compared to a 84\% success rate in simulation. 

These results demonstrate that obstacle avoidance using solely a multiranger is challenging, as drift and limited visibility are the most prominent causes for crashing. In most crashes, the drone would either not see the obstacle or keep rotating close to an obstacle and eventually crash into the obstacle due to drift. The baseline serves to put our algorithm into perspective, i.e., it shows the robustness of our obstacle avoidance and source seeking strategy. A deteriorated success rate in the `HIGH OBS' scenario is expected, as it has never seen such a scenario in simulation. \textit{Despite some loss in success rate when adding more obstacles, our approach shows greater resilience to task complexity when compared to the baselines. It generalizes better than other baselines, even beyond the simulated environments, showing tinyRL's potential.}

\begin{figure} 
\adjustbox{valign=b}{
\begin{minipage}[b]{\linewidth}
    \centering
  \subfloat[ Normalized light measurements $c$ and its low-pass version $c_f$.  \label{2a}]{%
       \includegraphics[width=0.5\linewidth]{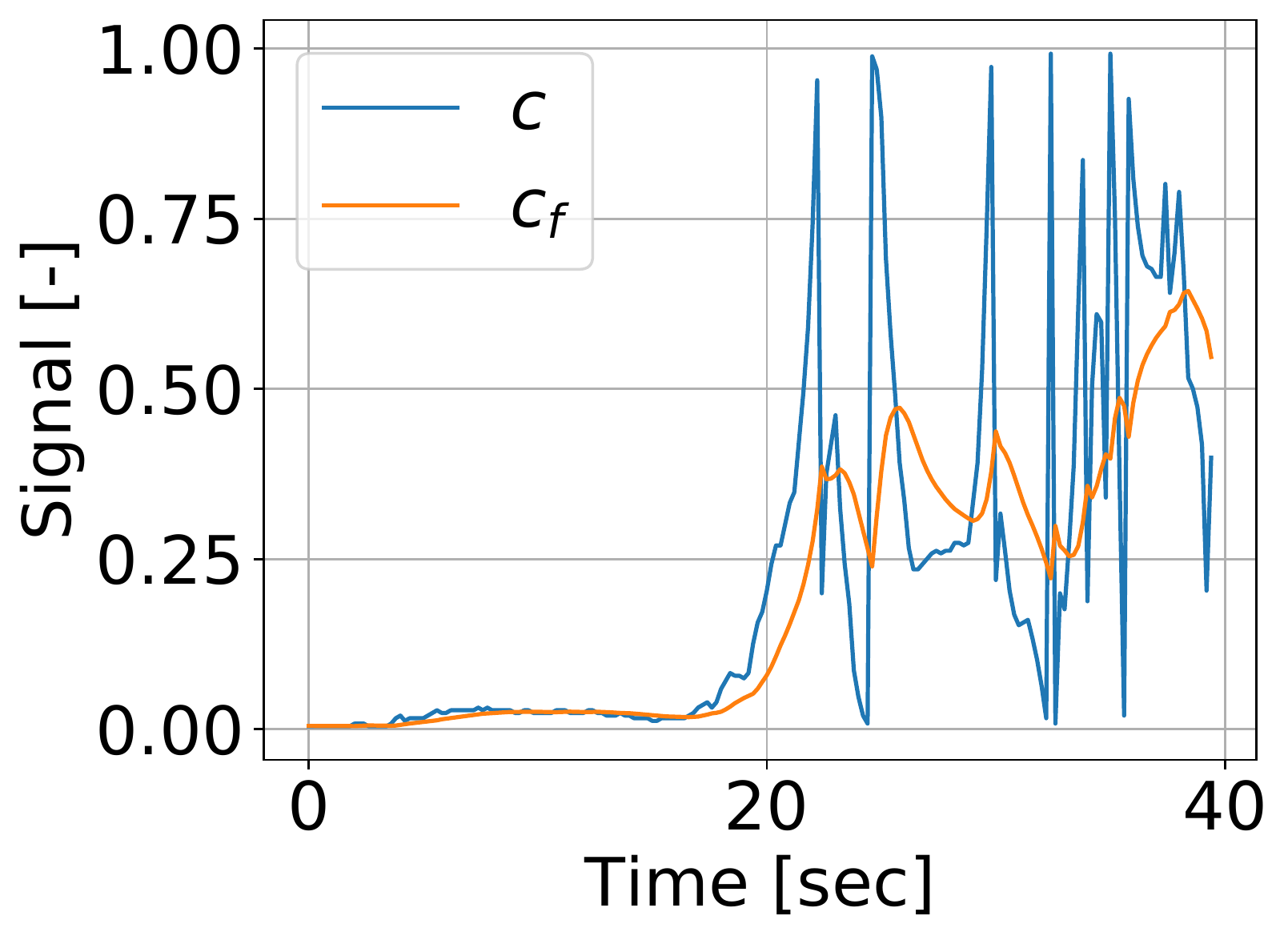}}
    \hfill
  \subfloat[ Policy inputs $s_1$ and $s_2$, as described in Equations~\ref{eq:s1} and~\ref{eq:s2}.\label{2b}]{%
        \includegraphics[width=0.48\linewidth]{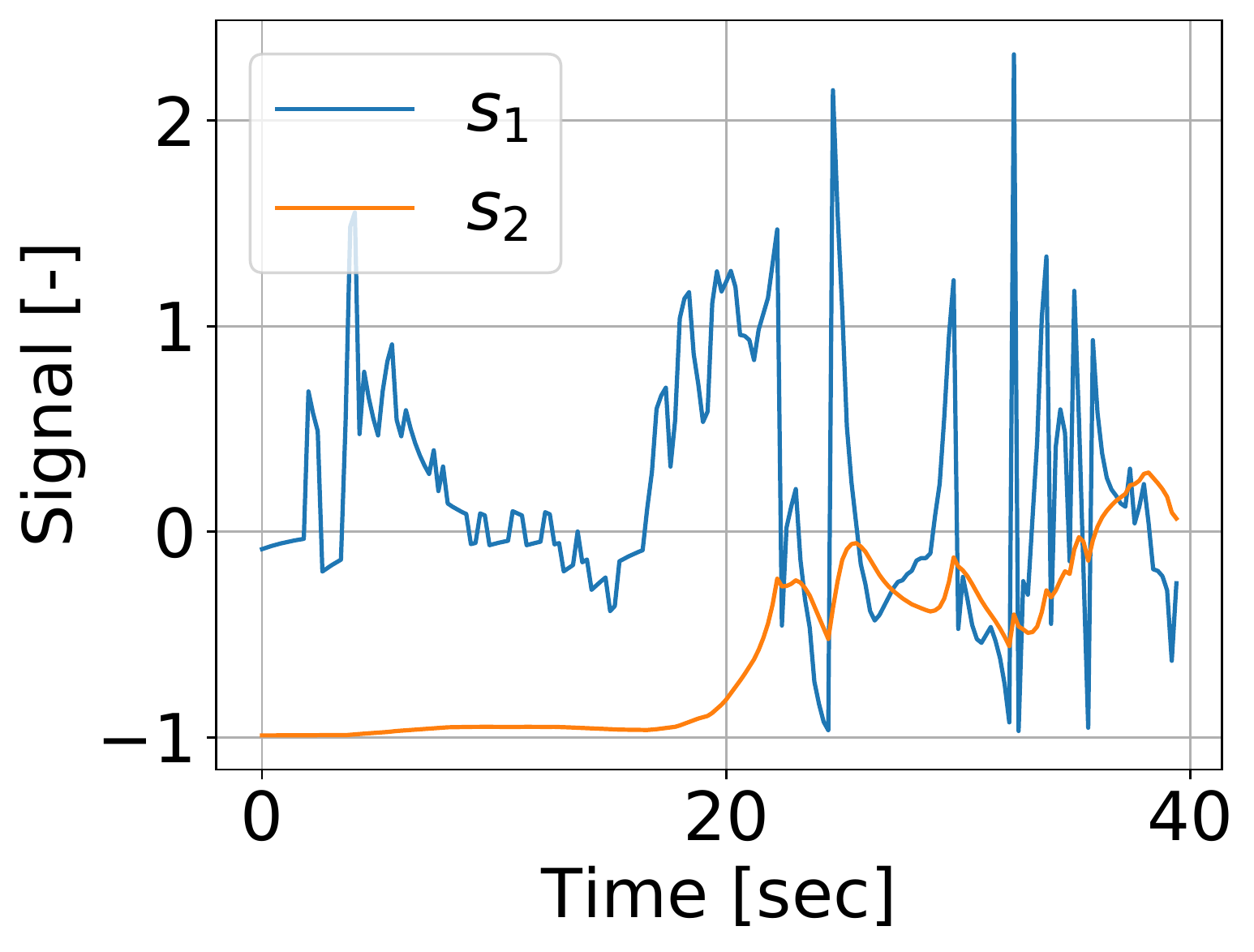}}
    \caption{Source measurements on the \nanodrone while seeking the source. The first 20 seconds the sensor provides little information.}
  \label{fig:source_traces}
  \end{minipage}%
  }
  \hfill
  \adjustbox{valign=b}{
\begin{minipage}[b]{\linewidth}
    \centering
    \includegraphics[width=\linewidth]{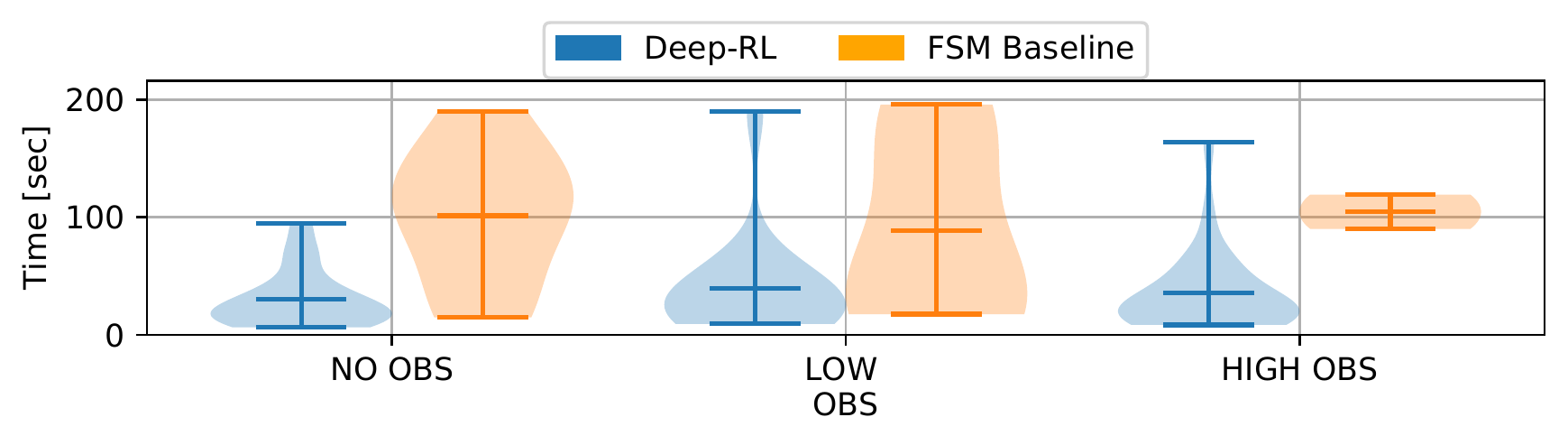}
    \caption{Mission time in success over 104 flight experiments, comparing our deep RL approach with the FSM baseline. }
    \label{fig:flight_exps_time}
\end{minipage}
}
\end{figure}

Our objective is not only to perform successful obstacle avoidance, but also to find the source in as little time as possible. Nano drones are characterized by their limited battery life. Therefore, efficient flight is an important metric when evaluating the viability of an approach for real applications. Figure~\ref{fig:flight_exps_time} shows the distribution of the mission time of successful runs, demonstrating an impressive advantage for our algorithm. Across the  obstacle densities, from low to high, our policy was 70\%, 55\%, and 66\% faster, respectively.

The baseline is again used to put our model into perspective. As demonstrated in~\cite{baseline}, the FSM is effective in exploring an area without any source information. Because of its random character, the baseline shows a more even distribution of mission times. 
The deep reinforcement learning approach has a small number of outliers with high mission time, often caused by the agent getting stuck in a certain trajectory in the dark. As shown in Figure~\ref{fig:source_traces}, the light gradient is limited far away from the source. The presence of noise makes it extremely hard for the agent to retrieve source information, at a great distance. We often observed a more direct path in the last \SI{2.5}\meter, compared to the initial exploration. \textit{Our approach performs robust obstacle avoidance and efficient source seeking using only a tiny neural network. }

\subsection{Behavior Analysis}
\label{sec:bahavior}


To better understand the behavior of the \nanodrone, we record source measurements and network inputs during flight. It can be seen that when far away from the source, extremely little information is present. In the first 20 seconds, almost no gradient is visible, forcing the agent to explore. The raw light readings are extremely noisy and unstable due to sensor noise, attitude changes and shadows.  Once the agent 'sees' the source, it keeps traveling up the gradient and doesn't go back to the dark zone. Finally, the features work as imagined, $s_1$ is a normalized light gradient with a low-pass filter and $s_2$ is a transformation of $c_f$. 



\subsection{Endurance and Power comparison}
\label{sec:endurance}
We consider endurance as a performance metric for our solution. By performing source seeking on the CrazyFlie, we add weight and CPU cycles. We determine endurance in hover and compare a stock hovering CrazyFlie with our source seeking solution. We swap the multiranger deck with light sensor for the battery holder, lowering the weight from \SI{33.4}\gram~to \SI{31.7}\gram~(\SI{-1.7}\gram ). The endurance observed with the stock CrazyFlie is 7:06.3, while our solution hovers for 6:30.8, reducing endurance by \SI{35.5}\second. With a battery capacity of 0.925 \texttt{Wh}, the average power consumption increased from \SI{7.81}\watt~to \SI{8.52}\watt (+\SI{0.71} \watt). It is expected that the vast majority of the extra power consumption comes from the extra weight, as the maximum consumption of the Cortex-M4 is \SI{0.14}\watt.

To put these numbers into perspective, we compare them to a CrazyFlie with a camera and additional compute~\cite{dronenet}. As shown in~\cite{dronenet}, endurance is reduced by \SI{100} \second~when adding the PULP-Shield, almost 3X more than in our experiments---\emph{the state of the art methods for vision-based navigation on nano drones have large impact on endurance, and hence different sensors are worth investigating. TinyRL applications will likely  make more use of sensors outside of the camera.}





\section{Discussion and Conclusion}
\label{sec:conclusion}


We show that deep reinforcement learning can be used to enable autonomous source seeking applications on nano drones, using only general-purpose, cheap, commodity MCUs. We trained a deep reinforcement learning policy that is robust to the noise present in the real world and generalizes outside of the simulation environment. We believe our tinyRL methodology is useful in other real-world applications too, as robots learn to adapt to noise and other implicit information. As our policy was trained on a general point source model, we believe it will provide a high-performance solution for other sources, such as radiation. By simply swapping the sensor, the \nanodrone may readily be  deployed to seek other sources. Versatile software and hardware will be important when deploying these robots in the real world, while exploring unknown environments and seeking an unknown source.

\bibliographystyle{IEEEtran}
\bibliography{example}  

\end{document}